\documentclass{ifacconf}
\usepackage{subcaption}
\usepackage{float}
\usepackage{amsmath}
\usepackage{amssymb}

\usepackage{algorithmic}
\usepackage{float}
\usepackage{graphicx}      % include this line if your document contains figures
\usepackage{natbib}        % required for bibliography

\begin{document}
\begin{frontmatter}

\title{Adaptive sampling using variational autoencoder and reinforcement learning} 

%\thanks[footnoteinfo]{Sponsor and financial support acknowledgment goes here. Paper titles should be written in uppercase and lowercase letters, not all uppercase.}

\author[First,Second]{Adil Rasheed} 
\author[First]{Mikael Aleksander Jansen Shahly} 
\author[Third]{Muhammad Faisal Aftab} 

\address[First]{Department of Engineering Cybernetics, Norwegian University of Science and Technology, Norway (e-mail: adil.rasheed@ntnu.no).}
\address[Second]{Mathematics and Cybernetics, SINTEF Digital, Norway.}
\address[Third]{Department of Engineering Sciences, University of Agder, Norway}

\begin{abstract}
Compressed sensing enables sparse sampling but relies on generic bases and random measurements, limiting efficiency and reconstruction quality. Optimal sensor placement uses historcal data to design tailored sampling patterns, yet its fixed, linear bases cannot adapt to nonlinear or sample-specific variations. Generative model-based compressed sensing improves reconstruction using deep generative priors but still employs suboptimal random sampling. We propose an adaptive sparse sensing framework that couples a variational autoencoder prior with reinforcement learning to select measurements sequentially. Experiments show that this approach outperforms CS, OSP, and Generative model-based reconstruction from sparse measurements.
\end{abstract}

\begin{keyword}
Adaptive Sampling, Variational Autoencoder, Reinforcement Learning
\end{keyword}

\end{frontmatter}
%========================================================

\section{Introduction}
Modern sensing systems increasingly rely on sparse measurements to reduce acquisition time, energy consumption, and bandwidth. Classical \emph{compressed sensing} (CS) provides a powerful framework for sparse sampling by exploiting signal sparsity in a generic transform basis and recovering the full signal via $\ell_1$-regularized optimization. However, because CS does not incorporate prior knowledge from training data and instead assumes sparsity in universal bases, the resulting reconstruction procedures are computationally expensive and often suboptimal for domain-specific signals \citep{Donoho2006CS}. 

An alternative is \emph{optimal sensor placement} (OSP), where information from historical data is used to tailor a problem-specific linear basis, typically via Proper Orthogonal Decomposition (POD) — and to select the most informative measurement locations through methods such as QR pivoting \citep{Manohar2018PODQR}. While OSP leverages prior structure more effectively than CS, it inherits the limitations of linear subspace models: the resulting sensors are globally defined and unable to adapt to nonlinear or sample-specific variations in new test signals. 

A different perspective was introduced by \cite{Bora2017CompressedSensingGM}, who showed that deep generative models can serve as powerful nonlinear priors for CS, enabling accurate reconstructions from randomly chosen sparse measurements. Although generative priors dramatically improve reconstruction quality, their measurement process remains agnostic to the content of the specific signal being sampled. Consequently, the randomly selected sampling locations are often suboptimal, leaving significant performance gains unrealized.

To address these limitations, we propose a new adaptive sparse sensing framework that combines generative priors with reinforcement learning (RL). By using a learned nonlinear manifold for reconstruction and a sequentially optimized sensing policy, our method adaptively selects the most informative measurements for each individual signal, surpassing both fixed OSP strategies and random CS sampling.

Section~\ref{sec:theory} introduces the theoretical background, including the problem setup, generative priors, and adaptive sampling. It formulates adaptive measurement selection as a RL problem and derives policy optimization strategies. Section~\ref{sec:methodology} describes the methodology, including dataset preparation, VAE training, and RL setup. Section~\ref{sec:resultsanddiscussions} presents the results and discusses them, followed by concluding remarks in Section~\ref{sec:conclusionandfuturework}.  

\section{Theory}
\label{sec:theory}
%%%%%%%%%%%%%%%%%%%%%%%%%%%%%%%%%%%%%%%%%%%%%%%%%%%%%%%%%
    \subsection{Compressed sensing} \label{subsec:CS}
%%%%%%%%%%%%%%%%%%%%%%%%%%%%%%%%%%%%%%%%%%%%%%%%%%%%%%%%%    
    Let $\boldsymbol{x} \in \mathbb{R}^n$ denote the unknown signal, and consider the linear measurement model
    \begin{equation}
        \boldsymbol{y} = \mathbf{C}\,\boldsymbol{x}, 
        \qquad 
        \mathbf{C} \in \mathbb{R}^{m \times n}, 
        \quad m \ll n .
    \label{eq:cs}
    \end{equation}
    Because the number of measurements is much smaller than the signal dimension, the system is underdetermined and the solution is not uniquely identifiable in general. Classical CS resolves this issue by exploiting \emph{sparsity}: if $\boldsymbol{x}$ has only a few nonzero entries and the measurement matrix $\mathbf{C}$ satisfies appropriate conditions (e.g.\ restricted isometry or incoherence), then $\boldsymbol{x}$ can be recovered by solving
    $$\min_{\boldsymbol{x}} \|\boldsymbol{x}\|_{0}
    \quad \text{subject to} \quad \|\boldsymbol{y} - \mathbf{C}\boldsymbol{x}\|_{2} < \epsilon .$$
    Since this $\ell_0$ problem is NP-hard, it is typically replaced by its convex relaxation \citep{Candes2006Stable}
    $$\min_{\boldsymbol{x}} \|\boldsymbol{x}\|_{1} \quad \text{subject to}
    \quad \|\boldsymbol{y} - \mathbf{C}\boldsymbol{x}\|_{2} < \epsilon ,$$
    which can be solved efficiently and still yields exact recovery under standard CS assumptions.
    
    In many applications, including imaging, the signal $\boldsymbol{x}$ may not be sparse in its natural domain but can be represented sparsely in a suitable transform basis. Let $\boldsymbol{\Psi}$ be such a basis (e.g.\ a Fourier or wavelet basis) and write
    $$\boldsymbol{x} = \boldsymbol{\Psi}\,\boldsymbol{s},$$
    where $\boldsymbol{s}$ is sparse. Substituting this expression into the measurement model leads to the equivalent optimization problem
    $$\min_{\boldsymbol{s}} \|\boldsymbol{s}\|_{1} \quad 
    \text{subject to} \quad 
    \|\boldsymbol{y} - \boldsymbol{\Phi}\,\boldsymbol{s}\|_{2} < \epsilon,
    \qquad \boldsymbol{\Phi} = \mathbf{C}\,\boldsymbol{\Psi}.$$
    
    Thus, CS enables accurate recovery from far fewer measurements than required by classical Nyquist sampling, provided that sparsity and appropriate measurement conditions are satisfied.

    %%%%%%%%%%%%%%%%%%%%%%%%%%%%%%%%%%%%%%%%%%%%%%%%%%%%%%%%%%%%%%%%%%%%%%%%
    \subsection{Optimal sensor placement via Proper Orthogonal Decomposition and QR decomposition} \label{subsec:ops}
    %%%%%%%%%%%%%%%%%%%%%%%%%%%%%%%%%%%%%%%%%%%%%%%%%%%%%%%%%%%%%%%%%%%%%%%%
    While CS provides a powerful framework for recovering high-dimensional signals from few measurements, solving the associated $\ell_1$ optimization problems can be computationally demanding, especially in large scale imaging applications. Moreover, CS relies on the assumption that the signal is sparse in some \emph{generic} basis (such as Fourier or Wavelets). When prior data or representative samples of the signal class are available, a more effective strategy is to replace these generic bases with a \emph{customized}, data-driven basis that captures the dominant structure of the signals, as proposed by \cite{Manohar2018PODQR}. In their work, they proposed the use of Proper Orthogonal Decomposition (POD). Instead of relying on a random sensing matrix $\mathbf{C}$ in Eq.\ref{eq:cs}, the aim is to \emph{design} it so that the chosen measurements provide maximal information for reconstructing $\boldsymbol{x}$ based on the available training data.
    
    Let
    $$\mathbf{X} =
    \begin{bmatrix}
    \boldsymbol{x}^{(1)} & \boldsymbol{x}^{(2)} & \cdots & \boldsymbol{x}^{(N)}
    \end{bmatrix}
    \in \mathbb{R}^{n \times N}$$
    be a collection of representative snapshots. Performing a singular value decomposition
    $$\mathbf{X} = \mathbf{U}\mathbf{\Sigma}\mathbf{V}^\top,$$
    we obtain the POD basis from the first $r$ left singular vectors,
    $$\mathbf{U}_r = 
    \begin{bmatrix}
    \boldsymbol{u}_1 & \cdots & \boldsymbol{u}_r
    \end{bmatrix}.$$
    Signals from the same class can be approximated efficiently as
    $$\boldsymbol{x} \approx \mathbf{U}_r \boldsymbol{a}, \qquad
    \boldsymbol{a} \in \mathbb{R}^r,$$
    So, the problem of reconstruction reduces to estimating the reduced coefficients $\boldsymbol{a}$.
    
    The sensor placement problem thus becomes one of selecting the measurement locations (rows of $\mathbf{C}$) that best preserve the information needed to infer $\boldsymbol{a}$. To achieve this, a QR decomposition with column pivoting is applied to the transpose of the POD basis:
    $$\mathbf{U}_r^\top \mathbf{P} = \mathbf{Q}\mathbf{R}.$$
    The permutation matrix $\mathbf{P}$ orders the spatial indices by their linear independence, and the first $m$ pivot columns correspond to the most informative sensor locations. The sensing matrix is then constructed as
    $$\mathbf{C} =[\boldsymbol{e}_{i_1}^\top \boldsymbol{e}_{i_2}^\top \cdots \boldsymbol{e}_{i_m}^\top]^{\top}$$
    where $\{i_1,\ldots,i_m\} =$ \text{pivot indices from QR}.
    
    This POD--QR framework provides a deterministic and computationally efficient alternative to $\ell_1$-based CS. By leveraging prior data to construct a tailored low-dimensional basis and selecting sensor locations via pivoted QR, the method produces highly informative measurement patterns and enables accurate reconstruction from only a small number of optimally placed sensors as shown in \cite{Menges2024rtp}.

    \subsection{Adaptive sampling with generative priors and RL}
    A limitation of POD-QR-based optimal sensor placement is that it produces a set of measurement locations derived from a linear subspace approximation of the training data, making it inflexible when test signals deviate from this structure or exhibit localized, sample-specific features. Moreover, because the POD bases are global and linear, the resulting sensors cannot adapt to nonlinear variations or uncertainties in individual reconstructions. To address these issues, we propose adaptive sampling with generative priors and RL. 
        \subsubsection{Generative Priors}

\begin{figure}
    \centering
    \includegraphics[width=\linewidth]{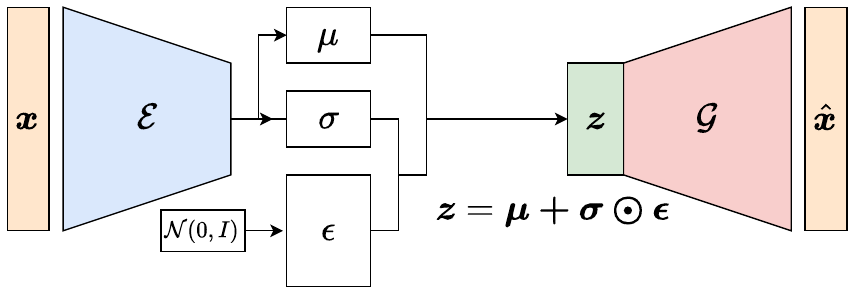}
    \caption{A VAE is trained offline on the original training dataset. After training, only the decoder ($\mathcal{G}$) is used when solving an optimization problem to estimate the latent code $\hat{\boldsymbol{z}}$.}
    \label{fig:VAE}
\end{figure}

We learn a generative prior by training a variational autoencoder (VAE) (Fig.\ref{fig:VAE}) on the original signals $\boldsymbol{x}$. 
The VAE assumes a Gaussian prior over the latent variables,
\begin{equation}
    p(\boldsymbol{z}) = \mathcal{N}(\boldsymbol{0}, \mathbf{I}),
\end{equation}
and is trained by minimizing the standard VAE loss:
\begin{equation}
\mathcal{L}_{\text{VAE}}
=
\underbrace{\| \boldsymbol{x} - \hat{\boldsymbol{x}} \|_2^2}_{\text{reconstruction}}
+
\underbrace{D_{\mathrm{KL}}\!\left(
    q_{\mathcal{E}}(\boldsymbol{z}\,|\,\boldsymbol{x})
    \,\|\, 
    \mathcal{N}(\boldsymbol{0}, \mathbf{I})
\right)}_{\text{Kullback Leibler regularization}} .
\end{equation}

\begin{equation}
\hat{\boldsymbol{x}} = \mathcal{G}(\boldsymbol{z}), 
\qquad 
\boldsymbol{z} \sim q_{\mathcal{E}}(\boldsymbol{z}\,|\,\boldsymbol{x}).
\end{equation}

Here, $\mathcal{E}$ denotes the encoder, which produces the approximate posterior 
$q_{\mathcal{E}}(\boldsymbol{z}\,|\,\boldsymbol{x})$, and $\mathcal{G}$ denotes the decoder.

After training, the decoder $\mathcal{G}$ serves as a learned generative prior, restricting reconstructed signals to lie on the manifold captured by the VAE. We assume that each signal can be approximated by a point on this generative manifold:
\begin{equation}
    \boldsymbol{x} \approx \mathcal{G}(\boldsymbol{z}^*),
    \qquad 
    \boldsymbol{z}^* \in \mathbb{R}^k,
\end{equation}
where the latent dimension satisfies $k \ll n$.

Reconstruction in downstream tasks reduces to solving for the latent vector that best explains the observation $\boldsymbol{y}$:
\begin{equation}
    \hat{\boldsymbol{z}}
    =
    \arg\min_{\boldsymbol{z} \in \mathbb{R}^k}
    \| \mathbf{C}\mathcal{G}(\boldsymbol{z}) - \boldsymbol{y} \|^2,
    \qquad
    \hat{\boldsymbol{x}} = \mathcal{G}(\hat{\boldsymbol{z}}).
\end{equation}

        \subsubsection{Adaptive Measurement Selection}
        
        We consider an adaptive setup where subsequent measurements are chosen sequentially based on the previously acquired data.
        \begin{figure}
            \centering
            \includegraphics[width=\linewidth]{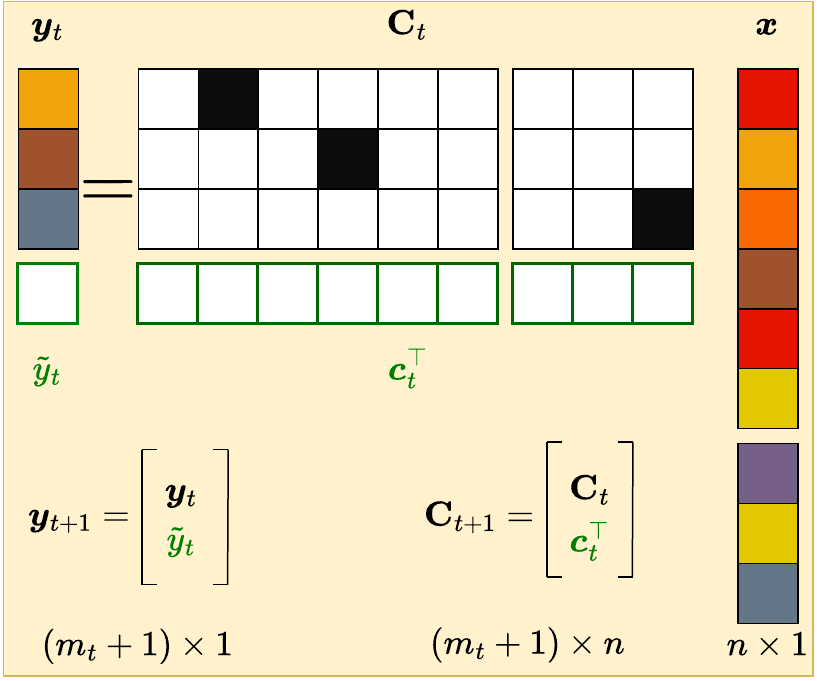}
            \caption{How the measurement matrix $\mathbf{C}$ the measurement vector $\boldsymbol{y}$ are expanded by the Reinforcement Learning agent}
            \label{fig:adaptivestage}
        \end{figure}
        At each step $t$ (see Fig.~\ref{fig:adaptivestage}), the adaptive sampler selects a new measurement vector
        \begin{equation}
            \boldsymbol{c}_t \in \mathbb{R}^n,
        \end{equation}
        which is typically restricted to a canonical basis vector. The corresponding scalar measurement is then
        \begin{equation}
            \tilde{y}_t = \boldsymbol{c}_t^\top \boldsymbol{x}, 
            \qquad y_t \in \mathbb{R}.
        \end{equation}
        
        We denote by $\mathbf{C}_t \in \mathbb{R}^{m_t \times n}$ and $\boldsymbol{y}_t \in \mathbb{R}^{m_t}$ the measurement matrix and measurement vector after $t$ adaptive steps. After incorporating the new measurement $(\boldsymbol{c}_t, y_t)$, we update:
        \begin{equation}
            \mathbf{C}_{t+1} = 
            \begin{bmatrix}
                \mathbf{C}_t \\
                \boldsymbol{c}_t^\top
            \end{bmatrix}
            \in \mathbb{R}^{(m_t+1) \times n},
            \qquad
            \boldsymbol{y}_{t+1} =
            \begin{bmatrix}
                \boldsymbol{y}_t \\
                \tilde{y}_t
            \end{bmatrix}
            \in \mathbb{R}^{m_t+1}.
        \end{equation}
        
        This recursive formulation ensures that, at each step, the measurement system grows by one row, with the goal of gradually refining the reconstruction of $\boldsymbol{x}$ using the most informative additional samples. The problem of selecting where to place sensors (i.e., which measurements to acquire) is formulated as an RL task. At each step, the agent observes the current reconstruction, decides where to measure next, and receives appropriate feedback. 
        
        \subsubsection{Action Space.}
        In our setup, each adaptive measurement corresponds to observing a single coordinate of $\boldsymbol{x}$. 
        Thus, the action space is the set of canonical basis vectors
        \begin{equation}
            \mathcal{A} = \{ \boldsymbol{c}_1, \boldsymbol{c}_2, \dots, \boldsymbol{c}_n \},
        \end{equation}
        where $\boldsymbol{c}_i \in \mathbb{R}^n$ has a 1 in the $i$-th position and zeros elsewhere. 
        Choosing $\boldsymbol{c}_i$ yields the scalar measurement
        \begin{equation}
            \tilde{y}_t = \boldsymbol{c}_i^\top \boldsymbol{x} = x_i.
        \end{equation}

        \subsubsection{State Representation.} 
        The state encodes all information available at step $t$. A minimal choice is the current set of probe $\mathbf{C}_t$ and adaptive measurements $\boldsymbol{y}_t$:
        \begin{equation}
            \boldsymbol{s}_t = (\mathbf{C}_t, \boldsymbol{y}_t).
        \end{equation}
            
        %In practice, the state is represented by a compressed embedding (e.g., an autoencoder or the probe measurements $\boldsymbol{y}_p$), so that the policy network can process it efficiently.
        
        \subsubsection{Transition Dynamics.}
        Taking action $\boldsymbol{c}_t$ results in observing the scalar measurement
        \begin{equation}
            \tilde{y}_t = \boldsymbol{c}_t^\top \boldsymbol{x},
        \end{equation}
        and updating the measurement set:
        \begin{equation}
            (\mathbf{C}_t, \boldsymbol{y}_t) \mapsto (\mathbf{C}_{t+1}, \boldsymbol{y}_{t+1}).
        \end{equation}
        
        \subsubsection{Rewards}
        The agent is provided a +1 reward if it results in reconstruction of the correctly classified images otherwise it is penalized with a -1 reward. For the classification of the reconstructed image a pretrained convolutional neural network (CNN) is used. 
        
        \subsubsection{Policy Optimization}
        The goal of the reinforcement learning agent is to maximize the total reward collected during one sensing episode. If we denote by $r_t$ the reward at step $t$, the total return from step $t$ is
        \begin{equation}
        R_t = \sum_{k=t}^{T} r_k ,
        \end{equation}
        where $T$ is the length of the episode. 
        Maximizing the expected return corresponds to improving reconstruction quality across episodes.  For each action $\boldsymbol{c}_t$ taken in state $\boldsymbol{s}_t$, the policy gradient is estimated as
        \begin{equation}
        \nabla_{\boldsymbol{\theta}} J(\boldsymbol{\theta}) \;\approx\; \sum_{t=1}^{T} \nabla_{\boldsymbol{\theta}} \log \pi_{\boldsymbol{\theta}}(\boldsymbol{c}_t \mid \boldsymbol{s}_t)\,(R_t - b_t),
        \end{equation}
        where $\pi_{\boldsymbol{\theta}}(\boldsymbol{c}_t \mid \boldsymbol{s}_t)$ is the probability of selecting action $\boldsymbol{c}_t$ under the current policy, and $b_t$ is a baseline (often chosen as the average return) to reduce the variance of updates.
        
        This gradient tells us how to change the policy network:
        \begin{equation}
        {\boldsymbol{\theta}} \;\leftarrow\; {\boldsymbol{\theta}} \;+\; \eta \, \nabla_{\boldsymbol{\theta}} J({\boldsymbol{\theta}}),
        \end{equation}
        where $\eta > 0$ is the learning rate. 
        Thus, the parameters ${\boldsymbol{\theta}}$ are nudged in the direction that makes high-return actions more probable and low-return actions less probable. By repeating this process over many episodes, the policy network gradually learns to favor measurement strategies that consistently improve signal reconstruction. Large updates of the policy parameters $\boldsymbol{\theta}$ can cause the policy $\pi_{\boldsymbol{\theta}}$ to collapse, for example by assigning almost all probability mass to a single action. To address this, \emph{Proximal Policy Optimization} (PPO) \citep{Schulman2017ProximalPO} introduces a clipped objective that prevents overly aggressive policy updates while retaining the simplicity of policy gradient methods. Let $\pi_{\boldsymbol{\theta}_{\text{old}}}$ denote the policy before the current update. 
        We define the probability ratio
        \begin{equation}
        r_t(\boldsymbol{\theta}) = 
        \frac{\pi_{\boldsymbol{\theta}}(\boldsymbol{c}_t \mid \boldsymbol{s}_t)}
             {\pi_{\boldsymbol{\theta}_{\text{old}}}(\boldsymbol{c}_t \mid \boldsymbol{s}_t)},
        \end{equation}
        which measures how the likelihood of action $\boldsymbol{c}_t$ has changed under the new parameters. 
        The PPO objective is then
        \begin{equation}
        L^{\text{PPO}}(\boldsymbol{\theta}) = 
        \mathbb{E}\!\left[
        \min\!\Big(
        r_t(\boldsymbol{\theta}) A_t,\;
        \operatorname{clip}(r_t(\boldsymbol{\theta}), 1-\epsilon, 1+\epsilon) A_t
        \Big)
        \right],
        \end{equation}
        where $A_t = R_t - b_t$ is the advantage (return relative to a baseline), and $\epsilon$ is a small constant (e.g.\ $0.1$ or $0.2$) controlling the maximum allowed policy deviation. The full algorithm is explained in Algorithm~\ref{appendix:algorithm}.

\section{Methodology}
\label{sec:methodology}
\subsection{Datasets}
The MNIST dataset \citep{deng2012tmd} is a widely used benchmark in machine learning, consisting of 70,000 grayscale images of handwritten digits from 0 to 9. Each image is  28×28 pixels and centered within the frame, providing a clean and standardized representation of individual digits. The dataset is divided into 50,000 training samples, and 10,000 validation, and test samples each, enabling consistent evaluation of classification and reconstruction models. Despite its simplicity, MNIST remains a valuable testbed for developing and validating new algorithms because it is small, easy to preprocess, and captures meaningful variations in handwriting style.
All pixel intensities are normalized to the range $[0,1]$ prior to training. The training subset is used to learn the generative prior via the VAE, while the test subset is reserved for evaluating reconstruction quality under the adaptive sensing policy. During RL training, random masking or undersampling is applied to emulate incomplete measurements, thereby allowing the RL agent to learn optimal sampling strategies.

\subsection{VAE training}
We use a convolutional VAE tailored to $28 \times 28$ grayscale MNIST digits. The encoder consists of three strided convolutional layers (with 1, 32, 64, and 128 channels and $4 \times 4$ kernels with stride~2 and ReLU activations), followed by a flattening layer and two fully connected layers that output the mean and log-variance of a latent vector of dimension $\boldsymbol{z}=64$. The decoder maps this latent representation back through a fully connected layer to a $4 \times 4 \times 128$ feature map and a symmetric stack of convolutional decoding layers, ending with a sigmoid activation to produce normalized images. The VAE is trained using the Adam optimizer \citep{Kingma2017aam} with a learning rate of $10^{-3}$, a batch size of 32, and 20 training epochs. The loss function combines a mean squared error reconstruction term with a KL divergence regularizer weighted by $\lambda = 10^{-5}$. As shown in Fig.~\ref{subfig:VAEloss}, the VAE was trained for 20 epochs, after which there was little improvement in performance on the validation set.

\subsection{RL training}
\begin{figure}%[htbp]
    \centering
    \begin{subfigure}{0.49\linewidth}
        \includegraphics[width=\linewidth]{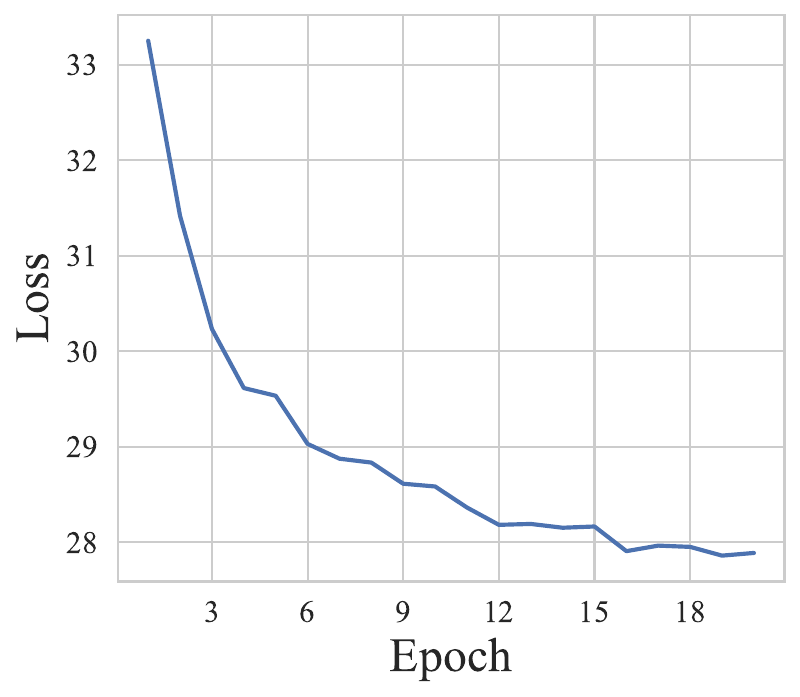}
        \caption{VAE Loss vs epochs}
        \label{subfig:VAEloss}
    \end{subfigure}
    \begin{subfigure}{0.49\linewidth}
        \includegraphics[width=\linewidth]{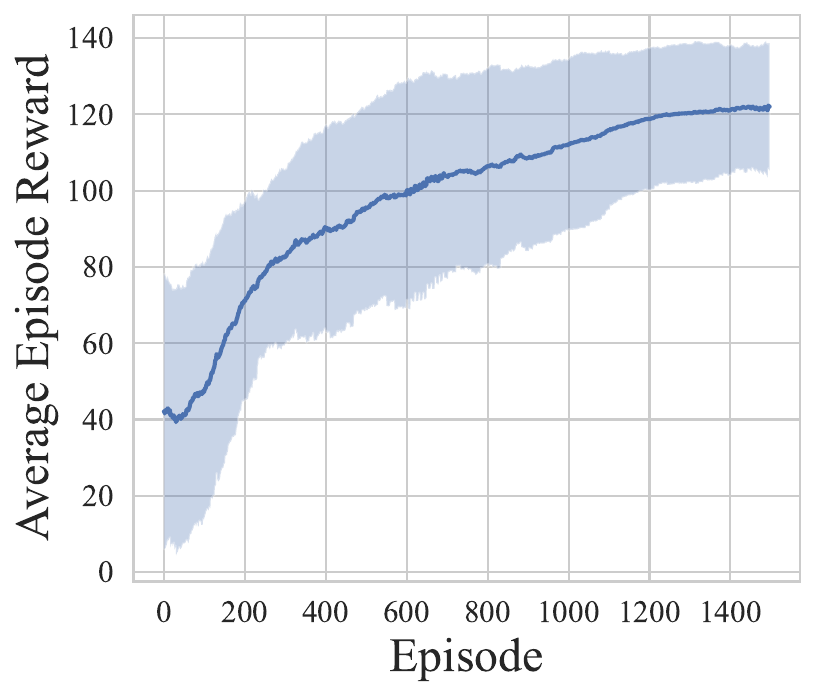}
        \caption{RL Reward vs episode}
        \label{subfig:RLreward}
    \end{subfigure}
    \caption{Evolution of the VAE and RL training}
    \label{fig:VAETrainingRLtraining}
\end{figure}
To learn an effective adaptive sampling strategy, we train a policy network that maps the current partial reconstruction into a distribution over all candidate measurement locations. The policy is implemented as a lightweight convolutional encoder composed of three sequential convolutional layers, Conv2d(C,32,3,1,1), Conv2d(32,64,3,2,1), Conv2d(64,64,3,2,1), each followed by a ReLU activation and a final flattening step that produces a compact feature embedding. On top of this representation, two linear heads output the action probabilities and value estimates required for policy optimization. Learning is performed end-to-end using PPO, which updates the policy through minibatched stochastic \emph{gradient ascent} on the clipped surrogate objective. During training, the agent receives the decoder-based reconstruction (mean of 20 ensembles) associated with the current measurement set, proposes the next sampling location, and collects rewards based on downstream classification performance. Optimization uses Adam with a learning rate of \(10^{-4}\), a minibatch size of 32, a discount factor \(\gamma = 0.99\), and a clipping parameter \(\epsilon = 0.2\). Each policy iteration runs several PPO epochs with an entropy regularization term that encourages exploration. Over many episodes, the agent converges toward sampling trajectories that consistently improve reconstruction quality under tight measurement budgets. Fig.~\ref{subfig:RLreward} illustrates the learning progress of the RL agent, showing that the average episode reward steadily increases over training episodes, indicating that the policy becomes progressively better at selecting informative measurements. 
   
\subsection{Metrics for Reporting the Results}
We use two metrics to evaluate reconstruction quality: the mean squared error (MSE) and the structural similarity index (SSIM) \citep{wang2004iqa}.
The MSE quantifies pixelwise deviation between the reconstructed image $\hat{\boldsymbol{x}}$ and the ground truth $\boldsymbol{x}$, defined as
\begin{equation}
\mathrm{MSE} = \frac{1}{N} \sum_{i=1}^{N} \left| \hat{\boldsymbol{x}}_i - \boldsymbol{x}_i \right|^{2},
\end{equation}
where $N$ is the number of pixels. Lower values indicate more accurate reconstructions.
We also use SSIM, which measures perceptual similarity by comparing local luminance, contrast, and structure. For two image patches $\boldsymbol{x}$ and $\hat{\boldsymbol{x}}$, it is given by
\begin{equation}
\mathrm{SSIM}(\boldsymbol{x},\hat{\boldsymbol{x}}) =
\frac{(2\mu_{\boldsymbol{x}} \mu_{\hat{\boldsymbol{x}}} + C_1)(2\sigma_{\boldsymbol{x}\hat{\boldsymbol{x}}} + C_2)}
     {(\mu_{\boldsymbol{x}}^2 + \mu_{\hat{\boldsymbol{x}}}^2 + C_1)(\sigma_{\boldsymbol{x}}^2 + \sigma_{\hat{\boldsymbol{x}}}^2 + C_2)},
\end{equation}
where $\mu_{\boldsymbol{x}}, \mu_{\hat{\boldsymbol{x}}}$ are local means, $\sigma_{\boldsymbol{x}}^2, \sigma_{\hat{\boldsymbol{x}}}^2$ are variances, $\sigma_{\boldsymbol{x}\hat{\boldsymbol{x}}}$ is the covariance, and $C_1 = 6.5025$, $C_2 = 58.5225$. are stability constants. Higher SSIM values indicate better structural fidelity.
\section{Results and discussions}
\label{sec:resultsanddiscussions}

\subsection{Reconstruction from VAE}
\begin{figure}[H]
    \centering
    \includegraphics[width=\linewidth]{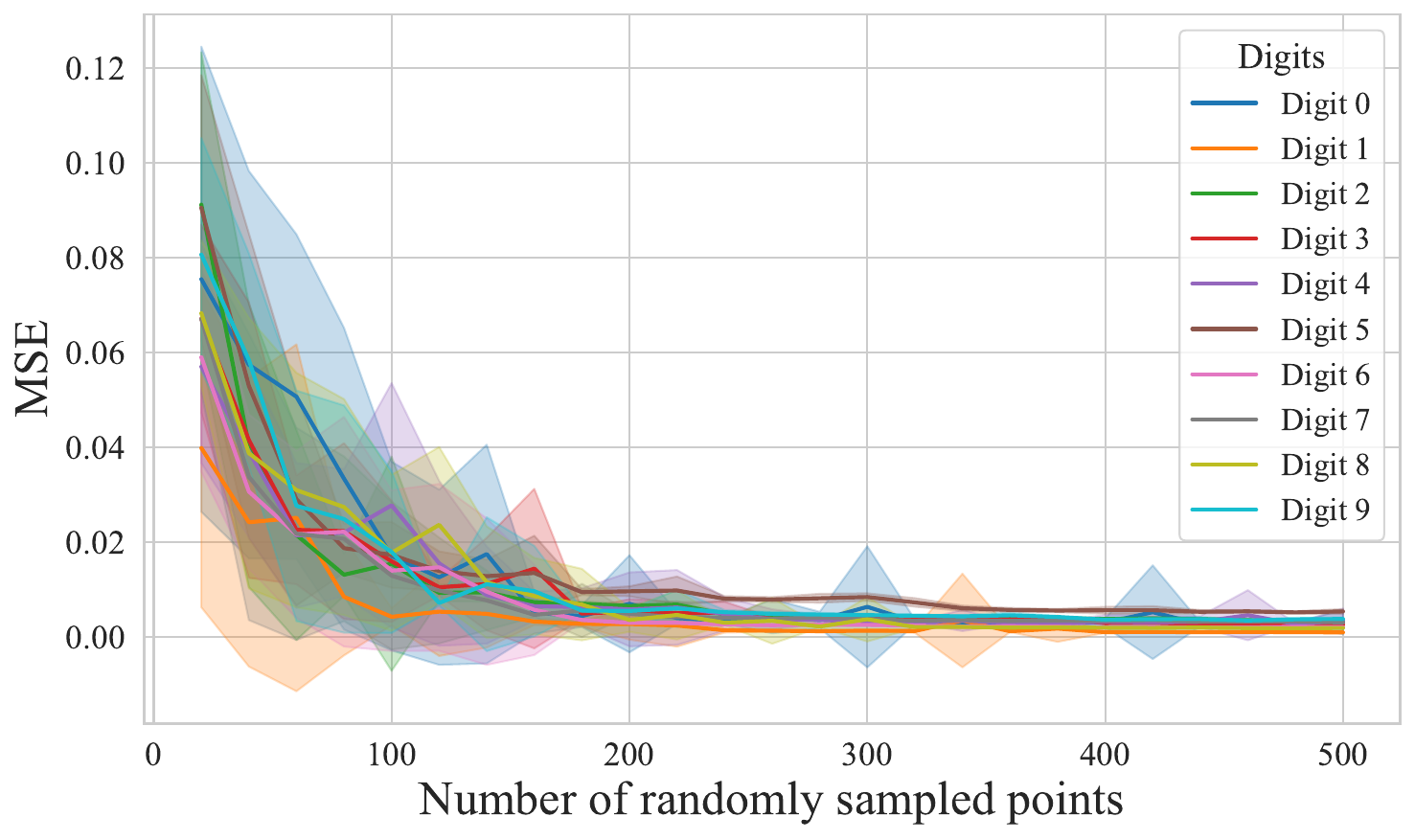}
    \caption{Reconstruction error from the trained VAE along with the uncertainty band for each digit}
    \label{fig:recon_error_vae}
\end{figure}

Fig. \ref{fig:recon_error_vae} illustrates how reconstruction accuracy improves as the number of sampled pixels increases for handwritten digits 0 through 9. Each colored curve corresponds to a different digit class and reports the MSE obtained from the trained VAE using randomly sampled points. 
\begin{figure}[H]
    \centering
    \includegraphics[width=\linewidth]{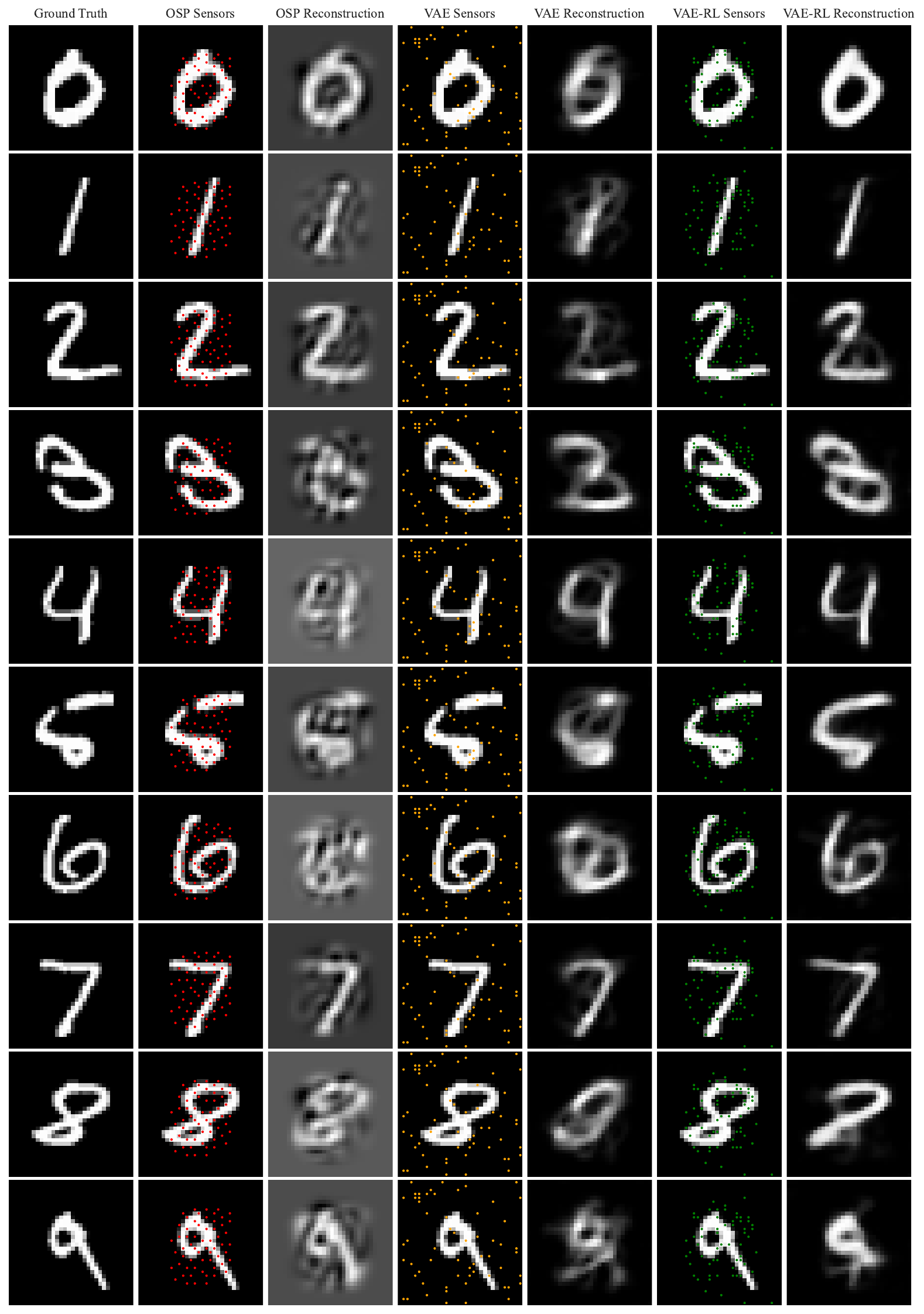}
    \caption{Comparision of reconstruction using OSP, VAE, and VAE-RL using 60 points}
    \label{fig:label8}
\end{figure}To quantify uncertainty, 20 independent VAE ensembles were evaluated for each digit, each ensemble using a different set of randomly selected measurement points. The shaded regions around each curve represent the variability across these ensembles, capturing the sensitivity of the reconstruction process to randomness in the sampled point locations. As expected, the reconstruction error decreases monotonically with more measurements, and the uncertainty bands narrow, indicating an increasingly stable and reliable recovery as the number of sampled points grows.

\subsection{Comparision of OSP, VAE and VAE-RL}
 
\begin{figure}
    \centering
    \begin{subfigure}{\linewidth}
        \includegraphics[width=1\linewidth]{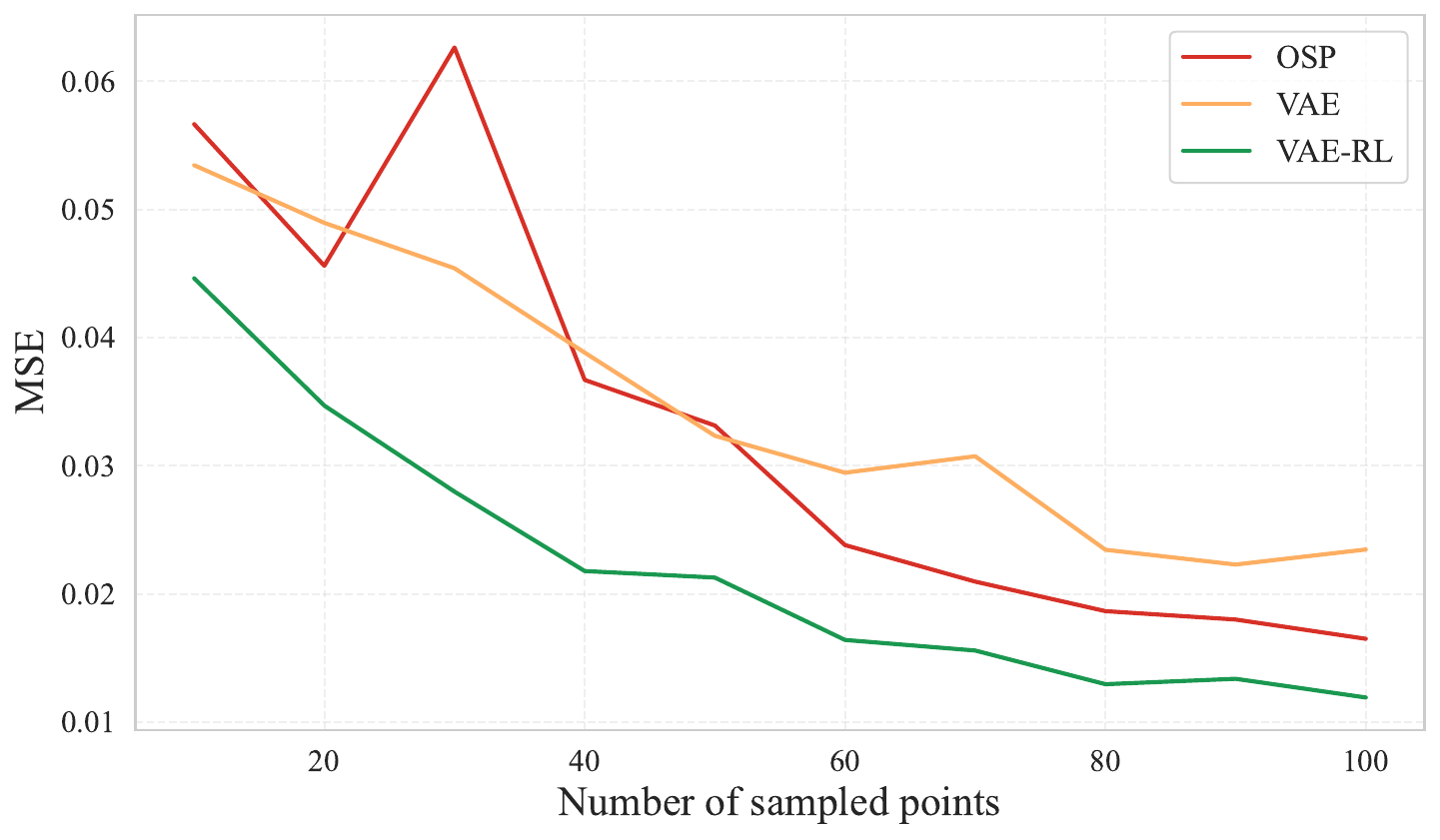}
        \caption{MSE (lower is better)}
        \label{subfig:mae_comparison}
    \end{subfigure}\\
    \begin{subfigure}{\linewidth}
        \includegraphics[width=1\linewidth]{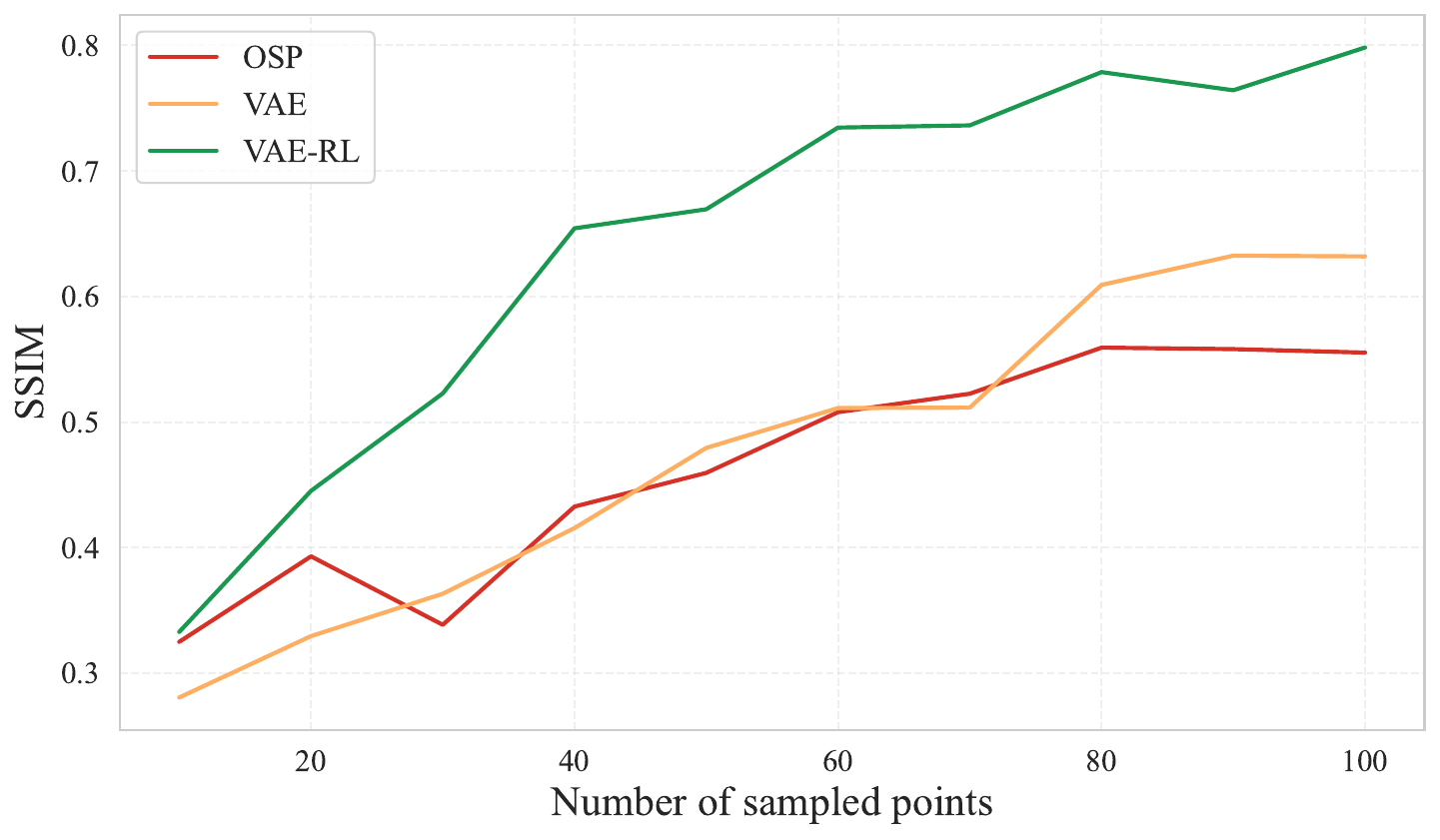}
        \caption{SSIM (higher is better)}
        \label{subfig:ssim_comparison}
    \end{subfigure}
    \caption{Average MAE and SSIR vs the number of sampled points}
    \label{fig:allLLMcomparison}
\end{figure}
The qualitative and quantitative results together (shown in Fig.~\ref{fig:label8} and Fig.~\ref{fig:allLLMcomparison}) illustrate clear differences between the three sensing strategies (OSP, VAE with random sampling, and the proposed VAE-RL adaptive sampling). The OSP approach, which relies on a fixed POD-QR sensor pattern, frequently places measurements in regions that do not align with the digit-specific strokes. As seen in the reconstruction grid, this leads to blurry or distorted reconstructions. This limitation is reflected quantitatively in the MSE plot, where OSP consistently produces higher reconstruction errors and lower SSIM values. The VAE with uniformly random sampling performs better, producing smoother and more globally consistent reconstructions. However, because the measurements are distributed blindly across the image, the method still struggles to capture local structures in more complex digits. In contrast, the VAE-RL strategy achieves the best performance, both visually and numerically. The RL agent adaptively concentrates sensors along the actual strokes of each digit, producing reconstructions with sharper edges and more accurate shapes. This adaptive behavior yields significantly lower reconstruction errors. Correspondingly, SSIM scores remain consistently high on average. These results demonstrate that adaptively allocating measurements to the most informative regions of each individual image enables substantially improved reconstruction fidelity compared to both fixed and randomly sampled sensing strategies.
    
\section{Conclusion and future work}
\label{sec:conclusionandfuturework}
This paper introduces an adaptive sparse sensing framework that integrates deep generative priors with reinforcement learning (RL) to overcome the limitations of classical compressed sensing (CS) and optimal sensor placement (OSP). While CS relies on generic bases and random sampling, VAE-based CS relies on random sampling, and OSP produces fixed linear measurement patterns that cannot adjust to nonlinear or sample-dependent variations, the proposed approach tailors each measurement to the specific signal being reconstructed. By combining a VAE priors with a RL policy learned through PPO, the method adaptively selects informative sampling locations and achieves significantly improved reconstruction accuracy compared to OSP, and VAE based CS with random measurements.

Several promising directions remain open. First, extending the framework to higher-resolution images and more complex datasets would test the scalability of both the VAE prior and the RL policy. Second, incorporating uncertainty quantification into the generative model or the policy could enable risk-aware measurement selection. Third, exploring alternative generative models such as diffusion models or Generative Adversarial Networks (GANs) may further improve reconstruction fidelity. Finally, applying adaptive sampling to real sensing hardware and dynamical systems represents an important step toward deploying these methods in practical imaging, robotics, and remote sensing applications.

\section*{DECLARATION OF GENERATIVE AI AND AI-ASSISTED TECHNOLOGIES IN THE WRITING PROCESS} During the preparation of this work the author(s) used ChatGPT-5 in order to improve the readability of this article.. After using this tool/service, the author(s) reviewed and edited the content as needed and take full responsibility for the content of the publication.

\bibliography{ifacconf}     
\section*{Appendix}
    \begin{algorithm}[H]
    \caption{Adaptive Compressed Sensing with Reinforcement Learning anf Generative Prior}
    \label{appendix:algorithm}
    \begin{algorithmic}[1]
    \REQUIRE Unknown signal $\boldsymbol{x} \in \mathbb{R}^n$, generative prior $\mathcal{G}(z)$, probe budget $m_p$, adaptive budget $m_a$
    \STATE \textbf{Probe stage:} Acquire $m_p$ random probe measurements $\boldsymbol{y}_p = \mathbf{C}_p \boldsymbol{x}$
    \STATE Initialize measurement set $(\mathbf{C}_0, \boldsymbol{y}_0) = (\mathbf{C}_p, \boldsymbol{y}_p)$
    \FOR{$t = 1$ to $m_a$}
        \STATE Encode state $\boldsymbol{s}_t$ from current $(\mathbf{C}_t, \boldsymbol{y}_t)$
        \STATE Sample action $\boldsymbol{c}_t \sim \pi_\theta(\cdot | \boldsymbol{s}_t)$
        \STATE Acquire measurement $\tilde{y}_t = \boldsymbol{c}_t^\top \boldsymbol{x}$
        \STATE Update measurement set $(\mathbf{C}_{t+1}, \boldsymbol{y}_{t+1}) = ([\mathbf{C}_t; \boldsymbol{c}_t^\top], [\boldsymbol{y}_t; \tilde{y}_t])$
        \STATE Reconstruct signal via latent optimization:
        \[
            \hat{\boldsymbol{z}}_{t+1} = \arg\min_{\boldsymbol{z}} \| \mathbf{C}_{t+1} \mathcal{G}(z) - \boldsymbol{y}_{t+1} \|^2, \quad
            \hat{\boldsymbol{x}}_{t+1} = \mathcal{G}(\hat{\boldsymbol{z}}_{t+1})
        \]
        \STATE Compute reward $r_t$
        \STATE Update policy parameters $\theta$ using REINFORCE:
        \[
            \nabla_\theta J(\theta) = \mathbb{E}_{\pi_\theta}\Big[\sum_t \nabla_\theta \log \pi_\theta(c_t|s_t) (R_t - b_t)\Big]
        \]
    \ENDFOR
    \RETURN Final reconstruction $\hat{\boldsymbol{x}}$
    \end{algorithmic}
    \end{algorithm}   
\end{document}